\documentclass{article}
\usepackage[utf8]{inputenc}
\usepackage{amsmath}
\usepackage{nomencl}
\makenomenclature
\usepackage{multicol}
\usepackage{float}
\usepackage[ruled,vlined,linesnumbered]{algorithm2e}
\SetKwFor{For}{for (}{) $\lbrace$}{$\rbrace$}
\usepackage[a4paper, total={6in, 8in}]{geometry}
\usepackage{hyperref}
\hypersetup{
    colorlinks=true,
    linkcolor=blue,
    filecolor=magenta,      
    urlcolor=cyan,
}
\usepackage{url}
\usepackage{float}
\usepackage{threeparttable}
\usepackage{booktabs}
\usepackage[normalem]{ulem}
\useunder{\uline}{\ul}{}
\usepackage{boldline}
\usepackage{multirow}
\usepackage{cite}
\usepackage{graphicx}
\usepackage{amsmath}
\usepackage{amssymb}
\usepackage{amsthm}
\theoremstyle{definition}
\newtheorem{definition}{Definition}
\theoremstyle{plain}
\newtheorem{theorem}{Theorem}

\newcommand{\EDiv}{\mathbin{\text{$\vcenter{\hbox{\textcircled{\footnotesize/}}}$}}}

\title{Algorithmic Solution for Systems of Linear Equations, in $\mathcal{O}(mn)$ time}
\author{Nikolaos P. Bakas
}
\date{}

\begin{document}

\maketitle

\begin{abstract}
We present a novel algorithm attaining excessively fast, the sought solution of linear systems of equations. The algorithm is short in its basic formulation and, by definition, vectorized, while the memory allocation demands are trivial, because, for each iteration, only one dimension of the given input matrix $\mathbf X$ is utilized. The execution time is very short compared with state-of-the-art methods, exhibiting $> \times 10^2$ speed-up and low memory allocation demands, especially for non-square Systems of Linear Equations, with ratio of equations versus features high (tall systems), or low (wide systems) accordingly. The accuracy is high and straightforwardly controlled, and the numerical results highlight the efficiency of the proposed algorithm, in terms of computation time, solution accuracy and memory demands. The paper also comprises a theoretical proof for the algorithmic convergence, and we extend the implementation of the proposed algorithmic rationale to feature selection tasks.

\end{abstract}

\section{Introduction}

The solution of a linear system appears in the vast majority of Linear Algebra operations \cite{suli2003introduction}, as well as related numerical methods, in statistical modelling, machine learning algorithms, numerical solution of differential equations, etc. These algorithms are essential for applications in almost any discipline involving computations, such as Engineering, Physics, Data Science, Finance, etc., among others \cite{mathews1992numerical}. The history of attempts to solve a Linear System is long, comprising the well-known Gaussian Elimination Algorithm for square systems \cite{grcar2011mathematicians,grcar2011ordinary}. A variety of types occur when formulating a linear system, such as systems with equal number of equations and unknowns (formulated with square input matrices), or systems with a few coefficients compared to the number of Equations (so called tall or underdetermined systems \cite{wang2010unique,donoho2006simplest,demmel1993improved}), where an exact solution does not occur and we try to identify the best possible solution in terms of residual errors, as well as wide (or overdetermined) systems \cite{williams1990overdetermined,barrodale1974solution,bartels1968numerical}, with more coefficients than equations, which have infinite solutions, and we try to identify one. These both are non-square systems. Accordingly, the input matrix can be dense \cite{poirier1998numerically,barrachina2008solving}, with all the elements non zeros, or sparse \cite{bai1996unified,bruckstein2008uniqueness}, with a few non zeros elements. Furthermore, the systems might have real or complex solutions, depending on the application and formulation of the particular problem \cite{behera2012new}, while the zero or not right part of the Equations classify a system to homogeneous (all right-part elements are equal to zero), or non-homogeneous \cite{funderlic1981solution}.

The purpose of this paper is to present a novel algorithm
for the solution of linear systems of any type. The algorithm was found to run in low compute time, while the memory demands are trivial, especially for tall and wide systems of Equations. In Section \ref{sec:alg}, we define the basic formulation of the algorithm, followed by a mathematical procedure (Section \ref{sec:deriv}) indicating the necessary logical steps to derive the algorithm. In Section \ref{sec:conv}, we prove that the algorithm converges to a solution with zero residual errors. In Section \ref{sec:interpr}, we explain the algorithm, by using standard Linear Algebra representations, followed by the parallel implementation of the algorithm in Section \ref{sec:par}. Finally, in Section \ref{sec:resu}, we demonstrate empirical results obtained from a large number of experimental systems solved by the proposed method.

\section{The Algorithm}
\label{sec:alg}
The origins as well as main application of the algorithm, regard the solution of Linear Systems for Regression purposes, hence we utilize the notation often used in such problems, without negating the generality of the method. Let 
\begin{equation}
    \mathbf X \times \mathbf a =\mathbf y,
    \label{eq:System}
\end{equation}
be a system of linear equations, with $\mathbf X$ the given input matrix with dimensions $(m \times n)$, $\mathbf y (m \times 1)$ the given output as a column vector, and $\mathbf a$, the $(n \times 1)$ sought vector of coefficients, satisfying Equation \ref{eq:System} in the best possible manner in terms of residual errors. By utilizing the following Algorithm \ref{al:HPSCD}, we can obtain the unknown $\mathbf a$, when knowing the matrix $\mathbf X$ and vector $\mathbf y$. Accordingly, $\mathbf X_j$ denotes the $j^{th}$ column of $\mathbf X$ matrix, and $a_j$ the $j^{th}$ element of the vector $\mathbf a$, with $j \in \{1,2,\dots,n\}$.
Instead of solving the entire system $\mathbf X \times \mathbf a =\mathbf y$, the proposed algorithmic formulation comprises two basic ideas, which suggest to solve:
\begin{enumerate}
    \item at each step $j$ for only one column of $\mathbf X$, $\mathbf X_j$,
    \item for the current's step errors $\mathbf e = \mathbf X \times \mathbf a - \mathbf y$ instead of the vector $\mathbf y$,
\end{enumerate}
and update the current weights $\mathbf a$ and errors $\mathbf e$, accordingly. Hence, we incrementally modify the $j^{th}$ coefficient of $\mathbf a$, $a_j$ such that it yields to the best possible reduction of the errors $\mathbf e$. If we repeat this for all the columns $j$, and for some round $N$, the algorithm converges to the sought solution $\mathbf a$. This is presented in detail in Algorithm \ref{al:HPSCD}, which is the basic formulation, that can easily be modified to control the solution error at each step and \textit{break}, if it is under a certain threshold, in order to control the trade-off among accuracy and execution time. Furthermore, more than one column of $\mathbf X$ could be utilized at each step, as we present in Section \ref{sec:par}, in order to parallelise the algorithm. Other variations of the basic algorithm can also be implemented; such instead of a serial execution of indices $j$, one could peak a randomly selected index $j \in \{1,2,\dots,n\}$ and check the corresponding convergence. $\mathbf 0_n$ denotes a vector with $n$ zeros.

We name the solver \textbf{HPSCD}, as it is a \textbf{H}igh \textbf{P}erformance \textbf{S}olver, based on \textbf{C}olumn-wise \textbf{D}ot products.

\begin{figure}[H]
\centering
\begin{minipage}{.7\linewidth}
\begin{algorithm}[H]
\SetAlgoLined
\KwData{$\mathbf X,\mathbf y$}
\KwResult{$\mathbf{a}$}
$\mathbf a = a_{j \in \{1,2,\dots,n\}} = \mathbf 0_n$ (initial guess)\\
$\mathbf e=\mathbf y - \mathbf X \times \mathbf a$\\
\For{$i \in \{1,2,\dots,N\}$}{
    \For{$j \in \{1,2,\dots,n\}$}{
    $\begin{aligned}
    da=\frac{\langle\,\mathbf X_j,\mathbf e \rangle}{\langle\,\mathbf X_j,\mathbf X_j\rangle}
    \end{aligned}$
    \\
    $\begin{aligned}
    \mathbf{e} \leftarrow \mathbf{e} - \mathbf X_j \times da
    \end{aligned}$
    \\
    $\begin{aligned}
    a_j  \leftarrow a_j + da
    \end{aligned}$
    }
}
\Return $\mathbf{a} \colon \mathbf X \times \mathbf{a} =\mathbf y$
\caption{The HPSCD Algorithm}
\label{al:HPSCD}
\end{algorithm}
\end{minipage}
\end{figure}

\section{Derivation of the Algorithm}
\label{sec:deriv}

Let $k$ be an iterator, increasing during each step of both the inner loop $\operatorname{\mathbf{for}} \; j \in \{1,2,\dots,n\}$, as well as the outer loop $\operatorname{\mathbf{for}} \; i \in \{1,2,\dots,N\}$:
\[
k \in \{1,2,\dots, N \times n\}.
\]

Accordingly, at any step $k$, we may write
\[\mathbf e_{k}=\mathbf y-\mathbf X \times \mathbf a_k,\]
where $\mathbf a_k$ are the coefficients at the current step $k$. Accordingly, for the next step $k+1$, we obtain that
\[\mathbf e_{k+1}=\mathbf y-\mathbf X \times (\mathbf a_k+\mathbf{da}_{k}),\]
where $\mathbf{da}_k$ denotes the change of the sought weights $\mathbf a$ at the step $k$. Hence
\[\mathbf e_{k+1}=\mathbf y-\mathbf X \times \mathbf a_k - \mathbf X \times \mathbf{da}_k\]
\begin{equation}
    \mathbf e_{k+1}=\mathbf e_{k} - \mathbf X \times \mathbf{da}_k
    \label{eq:ek1-ek}
\end{equation}

At each step $k$, we search for changes $\mathbf{da}_k$, relative to the current $\mathbf a_k$, such that after a number of iterations $k = N \times n$, the ultimate errors to be $\mathbf e_{k+1}\xrightarrow{}{} \mathbf 0_n$.

Thus, by setting $\mathbf e_{k+1} = \mathbf 0_n$, we obtain
\[\mathbf e_{k}=\mathbf X \mathbf{da}_k.\]

However, as aforementioned, we could perform this operation, not for the entire vector $\mathbf a$, but for a part of it. Accordingly, we select a single column of $\mathbf X$ with index $j$, as described in Algorithm \ref{al:HPSCD}, and we derive that
\begin{equation}
    \mathbf e_{k}=\mathbf X_j {da}.
    \label{eq:e-da}
\end{equation}

At each step $k$, the errors $\mathbf e_{k}$ and the column $\mathbf X_j$ of $\mathbf X$ are known.
Hence, we may solve for $da$, and we obtain
\begin{equation}
    {da}=(\mathbf X_j^T\mathbf X_j)^{-1} \mathbf X_j^T\mathbf e_k.
    \label{eq:da}
\end{equation}

$\mathbf X_j$ is a vector, thus $\mathbf X_j^T\mathbf X_j$is a number, hence we obtain
\[{da}=\frac{\mathbf X_j^T\mathbf e_k}{\mathbf X_j^T\mathbf X_j}\]
which is equivalent to
\begin{equation}
    {da}=\frac{\langle\,\mathbf X_j,\mathbf e_k \rangle}{\langle\,\mathbf X_j,\mathbf X_j\rangle}.
    \label{eq:da2}
\end{equation}

Henceforth, we formulate a vectorized representation of the computation of ${da}$ for each step $k$ for the current column $j$. 

Accordingly, at each step $j$, we compute the remaining residual errors $\mathbf e_{k+1}$, by

\[\mathbf e_{k+1}=\mathbf e_k-\mathbf X_j da,\]

and the updated weights $a_j$, by

\[a_j \xleftarrow{} a_j+da,\]

as per Equation \ref{eq:ek1-ek}, which also is a vectorised computation. 

Hence, by iterating for all $i$ and $j$, the Algorithm \ref{al:HPSCD} converges to the sought solution $\mathbf{a}$, with minimum errors
\[\mathbf e=\mathbf y-\mathbf X \mathbf{a}.\]

It is important to note that apart from the acceleration in the computations offered by utilizing vectors at each step (instead of the entire matrix $\mathbf X$), the necessary Memory Allocations are also trivial. This is also confirmed later in the numerical experiments Section \ref{sec:resu}. Accordingly, we may transfer the data to the GPU, and we can solve for large systems, as for each time, only a single column of $\mathbf X$ is required to be available on the memory. Hence, considering the limited memory available, we may solve systems on GPUs, which we could not do with standard algorithms.

\section{Algorithmic Convergence}
\label{sec:conv}

\begin{definition}[Minimum Errors]
We define the minimum possible errors
\begin{equation}
    e_{min} = \operatorname{min}_{\mathbf{a}} \| \mathbf y - \mathbf X \times \mathbf{a} \| ^2 \ge 0,
\end{equation}
for some $\mathbf{a}$ as the best possible solution of Equation \ref{eq:System}.

$e_{min}=0$, when the system has an exact solution (e.g. square systems), and $e_{min}>0$ otherwise (e.g. tall systems appearing in Linear Regression problems).

\end{definition}

\begin{theorem}
\emph{(HPSCD Convergence)}
\label{the:conv}
The HPSCD algorithm converges to an $\mathbf{a}$ solution of a system $\mathbf X \times \mathbf a =\mathbf y$, with minimum possible errors $e_{min}$.
\end{theorem}

\begin{proof}

As aforementioned, Equation \ref{eq:System} is not satisfied exactly for each step $k$, as it regards a least square, approximate solution. Thus instead of 
\[\mathbf e_k=\mathbf X_j {da},\]
we have
\begin{equation}
    \mathbf e_k=\mathbf X_j {da} +\mathbf{res}_k,
    \label{eq:res-basic}
\end{equation}

with $\mathbf{res}_k$ denoting the residual errors at each step $k$. Utilizing Equation \ref{eq:ek1-ek}, we may write
\begin{equation}
    \mathbf{res}_k=\mathbf e_{k+1}
    \label{eq:res}
\end{equation}

which should reach a minimum. Accordingly, by raising the errors of Equation \ref{eq:res-basic} to the power of $2$ element-wisely, the squared errors $\mathbf e_{k}^2$ are
\[\mathbf e_{k}^2=\mathbf X_j^2 \times {da}^2 +\mathbf{res}_{k}^2 +2 \times \mathbf X_j \times {da} \times \mathbf{res}_{k},\]

and, by utilizing the summation of the errors upon all $i \in \{1,2,\dots,m\}$, we obtain
\begin{equation}
    \sum_{i=1}^{m} \mathbf e_{k}^2=\sum_{i=1}^{m} \left( \mathbf X_j^2 \times {da}^2 +\mathbf{res}_{k}^2 +2 \times \mathbf X_j \times {da} \times \mathbf{res}_{k} \right).
    \label{eq:all-res}
\end{equation}

At each step $k$, we may say that we do a linear regression (see Section \ref{sec:interpr}) with $\mathbf e_k$ as the dependent variable and each $\mathbf X_j$ as a single predictor. Hence, we may easily show that
\begin{equation}
    \sum_{i=1}^{m} \mathbf X_j \times \mathbf{res}_k=0.
    \label{eq:res-x}
\end{equation}

Particularly, by multiplying Equation \ref{eq:res-basic} with $\mathbf X_j^T$, where we obtain
\[\mathbf X_j^T \times \mathbf{res}_k=\mathbf X_j^T \times \mathbf e_k - \mathbf X_j^T \times \mathbf X_j {da},\]

and using Equation \ref{eq:da}, we may write
\[\mathbf X_j^T \times \mathbf{res}_k=\mathbf X_j^T \times \mathbf e_k - \mathbf X_j^T \times \mathbf X_j \times (\mathbf X_j^T \times \mathbf X_j)^{-1} \times \mathbf X_j^T \times \mathbf e_k,\]

thus 
\[\mathbf X_j^T \times \mathbf{res_i}=\mathbf 0,\]
and Equation \ref{eq:res-x} holds true.

Henceforth, Equation \ref{eq:all-res} is written by
\[\sum_{i=1}^{m} \mathbf e_{k}^2=\sum_{i=1}^{m} \mathbf X_j^2 \times {da}^2 +\sum_{i=1}^{m} \mathbf{res}_{k}^2.\]

However, $da$ is non zero for at least one $j \in \{1,2,\dots,n\}$ otherwise all the columns $\mathbf X_j$ of $\mathbf X$ would be parallel as perpendicular to $\mathbf e$ (nominator of Equation \ref{eq:da2}), thus the errors 
\[\sum_{i=1}^{m} \mathbf e_{k}^2 > \sum_{i=1}^{m} \mathbf {res}_{k}^2,\]

and, by applying Equation \ref{eq:res}, we obtain that
\begin{equation}
    \sum_{i=1}^{m} \mathbf e_{k}^2 > \sum_{i=1}^{m} \mathbf e_{k+1}^2,
    \label{eq:decr}
\end{equation}
indicating that for each iteration $k+1$, the algorithm will always be exhibiting, explicitly decreased residuals than the previous step $k$, and hence 

\[\lim_{k\xrightarrow{}\infty}\sum_{i=1}^{m} \mathbf e_{k}^2 \xrightarrow{} e_{min},\]
and the Algorithm converges to a solution with residual errors equal to the minimum possible $e_{min}$. 

\end{proof}

\section{Time Complexity}

\begin{theorem}
\emph{(HPSCD Complexity)}
\label{the:compl}
The Time Complexity of the HPSCD algorithm is $\mathcal{O}(m n)$.
\end{theorem}

\begin{proof}

The algorithm \ref{al:HPSCD}, comprises an outer for-loop in $N$ iterations and an inner for-loop with three calculation steps, repeated $n$ times.

The computation of 
\[da=\frac{\langle\,\mathbf X_j,\mathbf e \rangle}{\langle\,\mathbf X_j,\mathbf X_j\rangle}\]

regards the dot product of $\mathbf X_j$ with $\mathbf e$, with time complexity $\mathcal{O}(m)$, the dot product of $\mathbf X_j$ with $\mathbf X_j$, with time complexity $\mathcal{O}(m)$, a division of two scalars with $\mathcal{O}(1)$, and a substitution of $da$ with $\mathcal{O}(1)$. 
Hence the total complexity of the first calculation step is $\mathcal{O}(2m+2)=\mathcal{O}(m)$.

\vspace{5mm}
The second step 
\[\mathbf{e} \leftarrow \mathbf{e} - \mathbf X_j \times da\]
\hspace{5mm}has a complexity of $\mathcal{O}(m+m+m)=\mathcal{O}(3m)=\mathcal{O}(m)$.

\vspace{5mm}
The third step 
\[a_j \leftarrow a_j + da\]
\hspace{5mm}has a complexity of $\mathcal{O}(1+1)$, for the addition and the substitution.

\vspace{5mm}
Hence, the total complexity of each iteration of the inner loop is $\mathcal{O}(m+m+1)=\mathcal{O}(m)$.

\vspace{5mm}
We repeat this $n$ times, hence the total complexity of the inner loop is $\mathcal{O}(m n)$.

\vspace{5mm}
From Theorem \ref{the:conv}, and Equation \ref{eq:decr} we deduce that the total solution error is being decreased with the running iterator $i \in \{1,2,\dots,N\}$. Hence, after finite iterations $N$, the algorithm will converge to any arbitrarily low error $\hat{e}_{min} \ge e_{min}$, in time $\mathcal{O}(m n)$.

\vspace{5mm}
Accordingly, we deduce that for any arbitrarily low error $\hat{e}_{min} \ge e_{min}$, the time complexity of the HPSCD algorithm is $\mathcal{O}(m n)$.
\end{proof}

It is important to note that the empirical evidence from the numerical experiments, confirms the linear relationship of compute time with $n$, and $m$.

\section{Interpretation}
\label{sec:interpr}
In order to explain further the rationale of the algorithm, we present below a series of logical steps with matrix notation. \\
Let 
\[\mathbf X \times \mathbf a =\mathbf y\]
a given system of equations, with 

\[\mathbf a = a_{j \in \{1,2,\dots,n\}}\]

the sought weights, that may be written in the form of

\begin{equation}
    \left[\begin{matrix}x_{11}&x_{12}&\cdots&x_{1j}&\ldots&x_{1,n}\\x_{21}&x_{22}&\cdots&x_{2j}&\ldots&x_{2,n}\\\vdots&\vdots&\ddots&\vdots&\ddots&\vdots\\x_{m,1}&x_{m,1}&\cdots&x_{m,j}&\ldots&x_{m,n}\\\end{matrix}\right]\times\left[\begin{matrix}a_1\\a_2\\\vdots\\a_j\\\vdots\\a_{n}\\\end{matrix}\right]=\left[\begin{matrix}y_1\\y_2\\\vdots\\y_{m}\\\end{matrix}\right]+\left[\begin{matrix}e_1\\e_2\\\vdots\\e_{m}\\\end{matrix}\right].
\end{equation}

The basic idea of the proposed algorithm is to solve at each step $k$
\begin{enumerate}
    \item for feature $\mathbf{X}_j$ instead of the entire matrix $\mathbf X$, and
    \item to solve for the current errors $\mathbf e$, instead of $\mathbf y$
\end{enumerate}

\begin{equation}
    \left[\begin{matrix}x_{1j}\\x_{2j}\\\vdots\\x_{m,j}\\\end{matrix}\right] \times da_j=\left[\begin{matrix}e_1\\e_2\\\vdots\\e_{m}\\\end{matrix}\right]
\end{equation}

Hence, we are actually updating one element of the sought weights $a_j$ at each step, in order to reduce the solution errors. Accordingly, in each step, the errors are decreased in the form of
\[
 \mathbf{e'}=\mathbf{e} - \mathbf X_j \times da,
 \]
and we update the $j$ th dimension of the weights $\mathbf a$ with the calculated $da$, and continue to the next step with the updated errors $\mathbf{e'}$. Hence, the algorithm is concise and simple, while it is vectorized straightforwardly. Accordingly, the supplementary memory allocation demands are meager.

\section{Parallel implementation of the Solver}
\label{sec:par}

The basic idea for the parallelization of the algorithm is to utilize at each step $k$ a sub-matrix of $\mathbf X$, $\mathbf X_{jj}$ for all rows and the particular columns 

\[jj \in \{k,k+1,\dots,k+thr-1\},\]

with $thr$ denoting the number of slicing columns. This approach can be performed serially as well. However, with multi-threading, we may seamlessly split the computation of the $da_{jj}$ to many threads $thr$. Although the error is not updated in the inner loop for each column $j$, for a batch of columns $jj$, it was found in the numerical experiments that if the $thr$ parameter is small with respect to the $n$ (e.g. $thr=\frac{1}{10} \times n$), the algorithm converges. Hence we may exploit the computational power of many threads when available.

\begin{figure}[H]
\centering
\begin{minipage}{.7\linewidth}
\begin{algorithm}[H]
\SetAlgoLined
\KwData{$\mathbf X,\mathbf y$}
\KwResult{$\mathbf{a}$}
$\mathbf a=\mathbf 0_n$ (or initial guess)\\
$\mathbf e=\mathbf y - \mathbf X \times \mathbf a$\\
\For{$i \in \{1,2,\dots,N\}$}{
    $\mathbf{aprev} \xleftarrow{} \mathbf{a}$\\
    \For{$j \in \{1,thr+1,2thr+1,\dots,n-thr+1\}$}{
        \For{$k \in \{j,j+1,\dots,j+thr-1\}$ \textbf{do in parallel}}{
        $\begin{aligned}
        a_k=a_k+\frac{\langle\,\mathbf X_k,\mathbf e \rangle}{\langle\,\mathbf X_k,\mathbf X_k\rangle}
        \end{aligned}$
        }
        $\begin{aligned}
        \mathbf{e}=\mathbf{e} - \mathbf X_{jj} \times (\mathbf{a}_{jj} -\mathbf{aprev}_{jj})
        \end{aligned}$
    }
}
\Return $\mathbf{a} \colon \mathbf X \mathbf{a} =\mathbf y$
\caption{The Parallel HPSCDP Solver}
\label{al:HPSCDP}
\end{algorithm}
\end{minipage}
\end{figure}
 
We see in Algorithm \ref{al:HPSCDP}, that we have a third for-loop, which can be naturally parallelized. The variable $a_k$ stands for the $k^{th}$ element of the sought weights $\mathbf a$, while in line 9 of the algorithm, we utilize the columns $\{j,j+1,\dots,j+thr-1\}$ of matrix $\mathbf X$ and corresponding elements of vectors $\mathbf{a}$, and $\mathbf{aprev}$ to update the errors $\mathbf{e}$.

\section{Feature Selection with HPSCD}
Although the original HPSCD algorithm is a solver for linear systems, we may modify it for feature selection tasks. The algorithm utilizes at each step $j$ only one feature of the input matrix $\mathbf X$, for the current errors $\mathbf e$. Instead of that, if we compute the $da$ using for all $j \in \{1,2,\dots,n\}$ the same errors $\mathbf e$ corresponding to the current step of the outer loop, we may retrieve information about which feature approximates best the current errors. 

In the following Algorithm \ref{al:HPSCDF}, we start with an empty array of the selected features $\mathbf J$, and we sequentially add features, in order to obtain the 

\[\mathbf J \subset \{1,2,\dots,n\},\]

comprising a number of features $F$, with minimum possible errors 
\[\sum_{i=1}^{m} \mathbf e^{2} = \sum_{i=1}^{m} y_i - pred_i\],

with

\[\mathbf pred = \mathbf X_{\mathbf J} \times \mathbf a.\]

Accordingly, in line 7 of Algorithm \ref{al:HPSCDF}, by utilizing the first step of the inner loop of Algorithm \ref{al:HPSCD}, we compute the $\mathbf da = da_j$, for each feature $j \in \{1,2,\dots,n\}$, and select the one that minimizes errors $\mathbf{E}_j$. Line 7 can be easily vectorized by using basic BLAS functions and hence, it is very fast. $\EDiv$ denotes the element-wise division of vectors, and $\operatorname {arg\,min}$ corresponds to the argument $j$, minimising $\mathbf e$. The matrix $\mathbf E$, contains all errors for the features $j$.

\begin{figure}[H]
\centering
\begin{minipage}{.7\linewidth}
\begin{algorithm}[H]
\SetAlgoLined
\KwData{$\mathbf X \vartriangleright$ all features' matrix, $\mathbf y \vartriangleright$ target variable} 
\KwResult{$\mathbf J \vartriangleright$ vector comprising the selected indices}
$\mathbf a=\mathbf 0_n$ (or initial guess)\\
$\mathbf e=\mathbf y - \mathbf X \times \mathbf a$\\
$\mathbf J = (\hspace{2mm})$\\
$\mathbf{E} = \mathbf 0_n$\\
$\mathbf{XTX} = \mathbf X^{T} \times \mathbf X$\\
\For{$f \in \{1,2,\dots,F\}$}{
    $\begin{aligned}
    \mathbf {da} = (\mathbf X^{T} \times \mathbf e) 
    \EDiv \mathbf{XTX}
    \end{aligned}$
    \\
    \For{$j \in \{1,2,\dots,n\{$}{
        $\begin{aligned}
        E_j \leftarrow \sum_{i=1}^{m} e_i -  X_{i,j} \times ( a_j + da_j)
        \end{aligned}$
    }
    $\begin{aligned}
    \hat j = \operatorname {arg\,min} \mathbf E
    \end{aligned}$
    \\
    $\begin{aligned}
    \mathbf J = (\mathbf J, \hspace{2mm} \hat{j})
    \end{aligned}$
    \\
    $\begin{aligned}
    a_{\hat{j}} \leftarrow \mathbf{da}_{\hat{j}}
    \end{aligned}$
    \\
    $\begin{aligned}
    \mathbf e \leftarrow \mathbf e - \mathbf X_{\hat j} \times a_{\hat{j}}
    \end{aligned}$
    \\
}
\Return $\mathbf J$
\caption{The HPSCDF Algorithm}
\label{al:HPSCDF}
\end{algorithm}
\end{minipage}
\end{figure}

Furthermore, instead of the outer for-loop, we may use a while-loop, with an unknown number of features $F$, and stop when the errors stop decreasing. This is particularly useful, when we do not know the optimal number of features, and can be further enhanced by using the cross-validation errors as a stopping criterion. 

Additionally, we may improve the process, by adding an outer loop, where we will remove features if, for the updated errors, a non-existing feature, yields lower errors than an existing one.

\section{Numerical Results}
\label{sec:resu}

In this section we present the numerical results in terms of accuracy, memory allocations, and computational time. The results run in Julia Language, and the comparison regards the corresponding wrappers for LAPACK and BLAS in Julia. The first 4 epxeriments were run on a machine with 6 threads and 16MB RAM, while results 5 to 12 run on a supercomputer with 200GB Ram and 80 cores. Accordingly, the first 4 benchmarks utilize 6 threads for BLAS computations, while benchmarks from 5-12 utilize 16 threads of BLAS, implemented in Julia \cite{bezanson2017julia} by calling the commands:

For the first 10 cases, the thr parameter of the parallel BAK \ref{al:HPSCDP} has the value of 50, and for the last two with big matrices $thr=1000$. All computations regard single float precision (Float32 in Julia), and were run ten times for each method, using the $@btime$ command in Julia and BenchmarkTools \cite{BenchmarkTools.jl-2016} package.

We can select the appropriate variation of the baseline algorithm, or optimize parameters such as $thr$, and accuracy threshold, especially if we have to solve multiple similar systems many times. Gaussian elimination and the corresponding Linear Algebra applications such as the computation of the Eigenvalues, were found faster than the proposed algorithm. However, if we want to get a fast solution of a system or initialize the weights, we may use the BAK algorithm for these cases as well.

\begin{figure}[H]   
\centering
\includegraphics[width=0.75\textwidth, keepaspectratio]{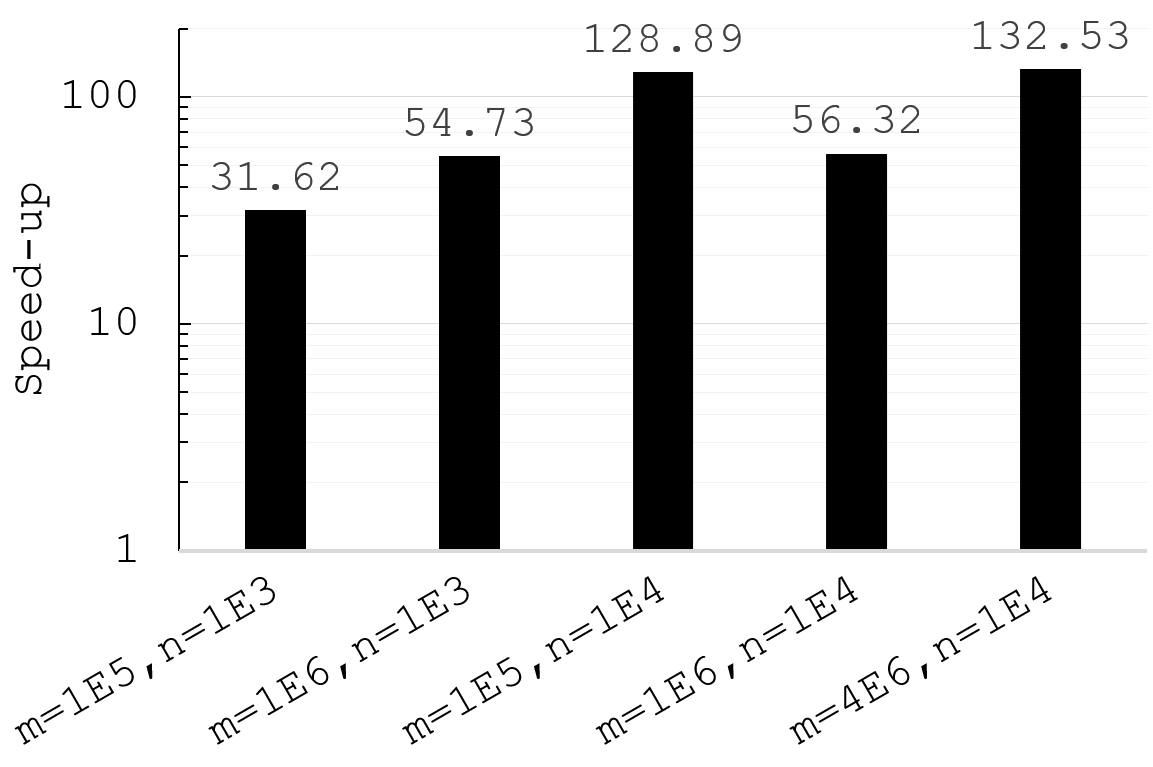}
\caption[]{Computing Time Speed-Up of solution time of HPSCD, versus standard BLAS solver.}
\label{fig:SpeedUp}
\end{figure}

\begin{figure}[H]   
\centering
\includegraphics[width=0.75\textwidth, keepaspectratio]{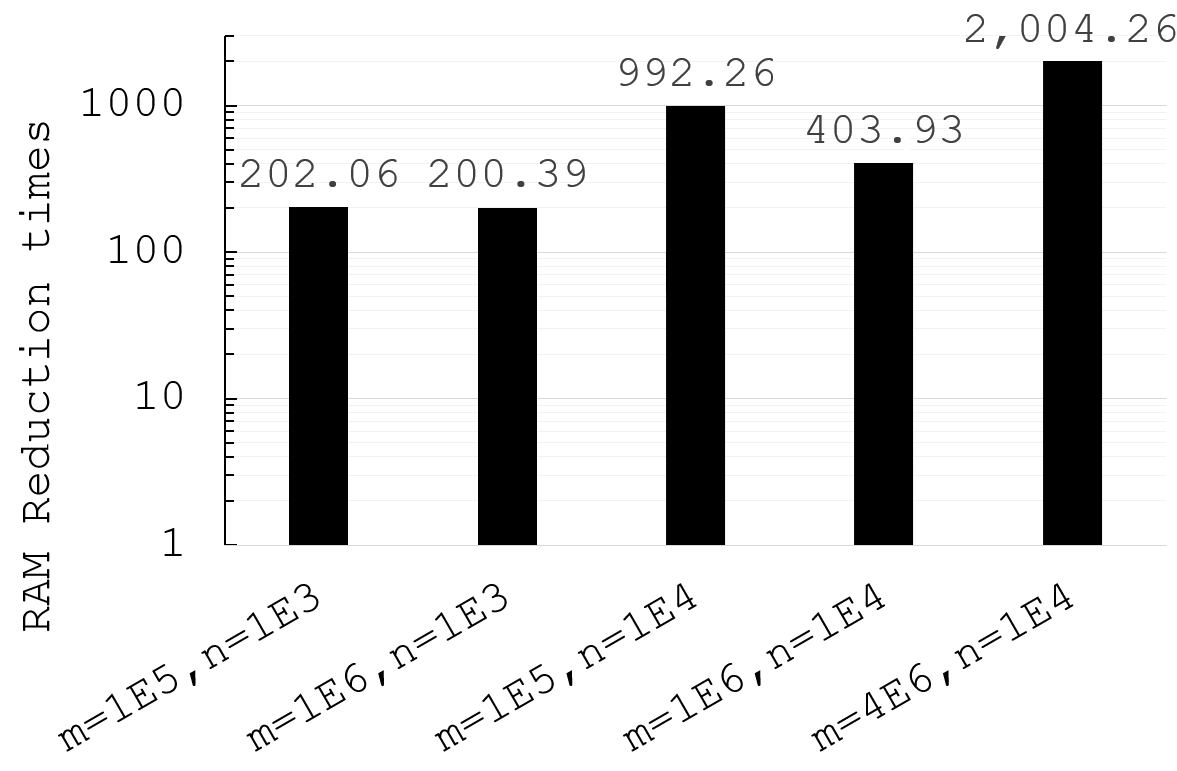}
\caption[]{Memory Reduction of HPSCD, versus standard BLAS solver.}
\label{fig:MemoryUp}
\end{figure}

\section{Conclusions}
\label{sec:concl}
The fast and accurate solution of a linear system of equations is a fundamental task for many Linear Algebra operations, as well as scientific and industrial applications. In this work we presented a novel algorithmic solution of Linear Systems, which exhibited high computational performance, especially for tall or wide, non-square systems of Linear Equations. The formulation of the basic algorithm is very simple, stemming for vector operations, which result in low demands in terms of memory allocation. By utilizing specific threshold apropos the accuracy of the algorithm within the iterative procedure, we can boost even more the performance, while the simplicity of the formulation, facilitates the implementation the algorithm at any programming Language, as well as modify it with respect to the particular problem. Accordingly, other formulations are also presented, for example the parallelization of the algorithm on many threads, as well as the execution on GPU accelerators. It was proven that the algorithm converges to the best possible solution, and the numerical experiments highlight the efficiency of the algorithm in terms of accuracy, speed, and memory allocations.

\nomenclature{$\mathbf 0_n$}{a vector with $n$ zeros}
\nomenclature{$N$}{the number of iterations}
\nomenclature{$n$}{the number of unknown coefficients}
\nomenclature{$m$}{the number of equations}
\nomenclature{$\mathbf X$}{input matrix with dimensions $m \times n$}
\nomenclature{$\mathbf y$}{vector with length $m$, comprising the right-hand values for each equation}
\nomenclature{$\mathbf a$}{the sought weights satisfying $\mathbf X \times \mathbf a = \mathbf y$}
\nomenclature{$\mathbf e$}{the solution errors $\mathbf e = \mathbf y - \mathbf X \times \mathbf a $}
\nomenclature{$max\_tol$}{threshold for the accuracy of the solution}
\nomenclature{$incr$}{number of columns sent to threads, for the case of the parallel implementation of the algorithm}
\nomenclature{$F$}{number of features to be selected}
\begin{multicols}{2}
\printnomenclature[1cm]
\end{multicols}

\bibliographystyle{IEEEtran}
\bibliography{refs}

\end{document}